\pdfoutput=1
\documentclass[10pt,letterpaper]{article}

\usepackage{makecell}
\usepackage[letterpaper,margin=1in]{geometry}
\usepackage{times}
\usepackage{amsmath,amssymb}
\usepackage{graphicx}
\usepackage{booktabs}
\usepackage{multirow}
\usepackage{xcolor}
\usepackage{xurl}
\usepackage[colorlinks=true,citecolor=blue!60!black,linkcolor=red!60!black,urlcolor=blue!60!black]{hyperref}
\hypersetup{pdftitle={Who Asks, Who Answers: Linear Attention Builds the Query in Hybrid Language Models},
  pdfauthor={Wenqi Zeng, Chen Feng, YuAng Chen}}
\usepackage{natbib}
\usepackage{caption}
\usepackage{subcaption}
\usepackage{float}
\usepackage{fancyhdr}

\newcommand{\ttt}[1]{\texttt{#1}}
\newcommand{\qsmall}{Qwen3.5-0.8B}
\newcommand{\qlarge}{Qwen3.5-9B}
\newcommand{\gsmall}{Granite4.0-1B}
\newcommand{\glarge}{Granite4.0-7B}

\title{Who Asks, Who Answers:\\
Linear Attention Builds the Query in Hybrid Language Models}

\author{%
  Wenqi Zeng\textsuperscript{1}\qquad
  Chen Feng\textsuperscript{2}\qquad
  YuAng Chen\textsuperscript{3}\\[6pt]
  \normalsize\textsuperscript{1}The Hong Kong University of Science and Technology\\
  \normalsize\textsuperscript{2}The Hong Kong Polytechnic University\\
  \normalsize\textsuperscript{3}The Chinese University of Hong Kong
}

\date{}

\begin{document}
\maketitle

\begin{abstract}
Hybrid language models interleave linear-attention layers with a small number of softmax-attention layers. Prior work largely attributes in-context retrieval to softmax attention, treating linear attention primarily as a mechanism for compressing past context. We revisit this division of labor and show that linear attention plays a direct role in retrieval by constructing the query used by softmax attention. Studying Qwen3.5 (Gated DeltaNet) and Granite4.0 (Mamba-2), we identify an Encode--Address circuit. An early Encode block preserves token identity, while an Address block constructs the final-position representation from which a softmax \emph{reader} forms its attention query. Transplanting this representation between prompts with different queried keys redirects retrieval in 0.97--0.99 of cases. The Address block's recurrent state carries the retrieval target, and its computation transfers across architectures: replacing Granite's Address block with Qwen's restores retrieval accuracy to 0.92--0.94 without updating either model's weights. Across six Qwen3.5 models, Address computation becomes distributed at larger scales over a chain of blocks, each guiding a successive reader, as the query no longer passes from one reader to the next through attention. These findings reveal a division of labor in hybrid retrieval: linear attention constructs the query, while softmax attention performs the lookup. Code is available at \url{https://github.com/yuang-chen/LinearQuery}.
\end{abstract}

\section{Introduction}

Hybrid language models combine linear-attention layers with a small number of softmax-attention layers. Linear attention, such as Gated DeltaNet \citep{yang2024gdn} or Mamba-2 \citep{dao2024mamba2}, compresses past context into a fixed-size state and scales linearly with sequence length, while softmax attention retains an explicit cache of past tokens. The combination aims to provide both the efficiency of recurrent state and the ability to retrieve exact information from earlier in the context, such as the value associated with a key introduced several paragraphs before. Recent open models, including Qwen3.5 and Granite-4.0-H, adopt this design with only one softmax layer for every few linear-attention layers \citep{qwen35, granite4}.

Prior work has shown that linear attention struggles with in-context retrieval \citep{jelassi2024repeat,arora2024zoology}, while softmax attention plays a central role in retrieval within hybrid models \citep{gna2025}. However, identifying where retrieval occurs does not explain how the surrounding linear-attention layers contribute to it. We therefore ask: \emph{What do linear-attention layers compute to support retrieval by softmax attention?}

We investigate this question through in-context key--value retrieval, where a model is given a dictionary (e.g., cherry=steel, mango=wood'') and must answer a question such as "What is the value of mango?'' with ``wood''. Although softmax attention can directly access the dictionary, we find that linear-attention layers are essential: removing the earliest linear-attention layer reduces retrieval accuracy to at most 0.05 across the models studied, even with all softmax layers intact.

To understand why, we first identify where retrieval emerges. Progressive layer pruning reveals a sharp increase in retrieval accuracy at a particular softmax layer, which we call the \emph{reader}. We then use two causal interventions on linear-attention blocks: \emph{removal} tests whether a block is necessary for retrieval, while \emph{transplantation} tests whether its output can redirect retrieval between prompts with different queried keys.

\begin{figure}[t]
    \centering
    \includegraphics[width=\textwidth]{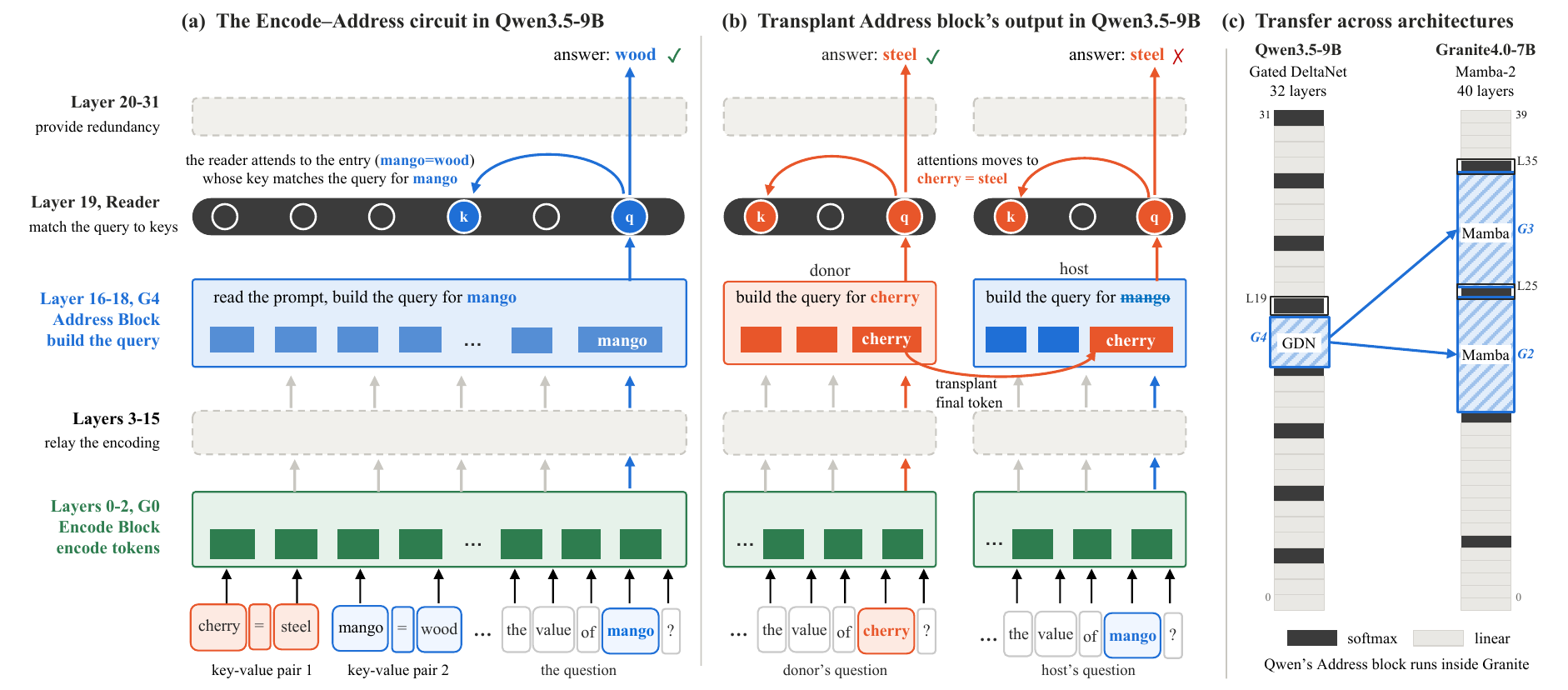}
    \caption{Overview. (a) The Encode block re-encodes token identity, the Address block constructs the retrieval query, and softmax attention retrieves the matching value. (b) Transplanting the address from prompt A into prompt B makes the model answer A's question. (c) Addresses transfer between Qwen3.5 (Gated DeltaNet) and Granite4.0 (Mamba-2)}
    \label{fig:overview}
\end{figure}

These interventions reveal an \emph{Encode--Address circuit} with two distinct roles (Fig.~\ref{fig:overview}a). The earliest linear-attention block, which we call \emph{Encode}, preserves token identity needed for retrieval. Its contribution to this task can be reproduced by a simple token-wise linear projection. The \emph{Address block}, immediately preceding the reader, constructs a representation at the final position that determines which dictionary entry the reader selects. Transplanting this output from one prompt into another redirects retrieval in 0.97--0.99 of cases in Qwen3.5-9B, even though the dictionary and question in the receiving prompt remain unchanged (Fig.~\ref{fig:overview}b).

We further investigate how the Address block performs this computation. Query--key interventions show that it primarily determines the reader's attention query rather than the dictionary keys. Transplanting its recurrent state also redirects retrieval. These results indicate that the Address block carries the retrieval target in its recurrent state and uses it to construct the representation from which the softmax reader forms its query.

Strikingly, the Address block's computation transfers across independently pretrained hybrid model families. Replacing the Address blocks of Granite-4.0-H (Mamba-2) with those from Qwen3.5-9B (Gated DeltaNet) restores retrieval accuracy to 0.94, without updating either model's weights (Figure~\ref{fig:overview}c). This suggests that the role of the Address block in guiding retrieval is a general property of hybrid language models, rather than one specific to a particular recurrent architecture. Across six Qwen3.5 models with growing sizes, we also find that Address computation becomes distributed at larger scales across a chain of linear-attention blocks, each guiding a successive softmax reader, because the query no longer passes from one reader to the next through attention.

Our contributions are:
\begin{enumerate}
\item \textbf{The Encode--Address circuit.} We identify two roles of linear attention in hybrid retrieval: an early Encode block supporting token identity and an Address block that determines which entry the softmax reader retrieves.
\item \textbf{The Address mechanism.} We show that the Address block uses its recurrent state to construct the representation that determines the reader's attention query, with upstream blocks providing redundant contributions.
\item \textbf{Cross-architecture transfer.} We demonstrate that the Address block can transfer between Gated DeltaNet and Mamba-2 without updating model weights, suggesting a shared retrieval mechanism across hybrid architectures.
\item \textbf{Scaling.} We show that larger models distribute Address computation over a chain of blocks, each guiding a successive reader, and trace this to how the query reaches the later readers. 
\end{enumerate}

\section{Related Work}
\label{sec:related}

\paragraph{Linear attention and recurrent sequence mixers.}
Transformers rely on softmax attention, which stores keys and values from previous tokens and attends over them explicitly at each step. While this provides direct access to the context, its key--value cache grows with sequence length. A broad class of efficient sequence models instead compresses past information into a fixed-size recurrent state, reducing the memory and computational cost of long-context processing. This includes linear-attention mechanisms such as fast weights and Gated DeltaNet \citep{schlag2021fastweights, yang2024gdn}, as well as state-space models such as Mamba-2 \citep{dao2024mamba2}. Although their update rules differ, these architectures share the computational property most relevant to our study: past information is maintained in a fixed-size state that is updated recurrently and used to produce the current token representation. For simplicity, we use \emph{linear attention} throughout the paper to refer to this broader class of fixed-state recurrent sequence mixers, including SSMs such as Mamba.

\paragraph{Hybrid architectures.}
Hybrid language models interleave linear-attention layers with a smaller number of softmax-attention layers \citep{lieber2024jamba, ren2024samba, waleffe2024empirical}. This design is now used by many released models, including Nemotron-H, Kimi Linear, Qwen3.5, and Granite-4.0-H \citep{nvidia2025nemotronh, kimi2025linear, qwen35, granite4}. Hybrids can also be obtained by distilling pretrained Transformers, retaining selected attention layers while converting the rest to recurrent layers \citep{wang2024mamballama}. Prior work mainly studies how many softmax layers to retain and where to place them. We instead ask what the surrounding linear-attention layers contribute to retrieval, which directly determines which of them a hybrid can remove, distill, or redesign without breaking retrieval.

\paragraph{Retrieval in recurrent and attention models.}
Recurrent models are known to struggle with copying and associative recall \citep{jelassi2024repeat, arora2024zoology, wen2024rnns}, a weakness commonly attributed to compression into a fixed-size state \citep{arora2024based}. Hybrid models mitigate this limitation by restoring a small number of softmax-attention layers, and \citet{waleffe2024empirical} show that even a small fraction of attention layers can substantially close the gap on recall-heavy tasks. \citet{gna2025} further trace this advantage to a small number of heads that implement a Gather-and-Aggregate mechanism, with hybrid models delegating aggregation to their attention layers. These works ask whether recurrent layers can perform retrieval. We instead study models in which retrieval already succeeds and ask how the computation is divided between recurrent and softmax layers. We find that linear attention is not bypassed: it constructs the query used by softmax attention.

\paragraph{Mechanisms of in-context retrieval.}
In Transformers, retrieval is often mediated by small sets of attention heads, including induction heads \citep{olsson2022induction}, heads that select answer letters in multiple-choice prompts \citep{lieberum2023mcq}, long-context retrieval heads \citep{wu2024retrieval}, and Gather-and-Aggregate heads \citep{gna2025}. These studies explain how attention heads read from context, but largely leave open how their queries are constructed. In the Gather-and-Aggregate framework, our \emph{Encode} block resembles a local gathering step and the reader performs aggregation, but the \emph{Address} block, which determines which entry the reader should retrieve, has no direct counterpart. 

Mechanistic studies of recurrent models have localized factual recall in Mamba \citep{sharma2024mamba, endy2025knockout} and shown that several interpretability methods developed for Transformers transfer to recurrent architectures \citep{paulo2024transfer}. To our knowledge, prior work has not disentangled the roles of linear and softmax layers within a hybrid retrieval circuit. We show that linear-attention layers serve two distinct functions: an early, generic re-encoding of token identity (\emph{Encode}) and a specific computation of the retrieval query (\emph{Address}) consumed by softmax attention.\looseness-1

\section{Case Study: In-Context Key--Value Retrieval}
\label{sec:case-study}

\subsection{Task and Model}
\label{sec:setup}
\paragraph{Retrieval task.}
We use a controlled in-context key--value retrieval task. Each prompt contains a dictionary of eight key--value pairs followed by a question about one key. The prompt is wrapped in each model's chat template, and the assistant response is prefilled so that the value is predicted as the next token:
\begin{quote}\small\ttfamily
Dictionary: elder=iron, carrot=red, cherry=steel, mango=wood, \dots\\
Question: What is the value of mango?\\
\textit{[assistant]} The value of mango is **
\end{quote}
Keys and values are single tokens, and each value has the same tokenization in the dictionary and in the answer. We call this format \ttt{chat8}. Two variants change the surface form while keeping the task, to test that our findings do not depend on one template. \ttt{list8} writes the dictionary as a bulleted lookup table with one \ttt{* key=value} entry per line, and rewords the question and answer prefix (``Which value does mango map to?'' and ``The entry mango maps to **''). \ttt{rev8} keeps the \ttt{chat8} question but states each entry with the value first (``iron is the value of elder''). Each format contains 100 prompts. \emph{Accuracy} is the fraction of prompts for which the queried value is the highest-logit token over the full vocabulary; all four models reach at least 0.99 on \ttt{chat8}.

We rely on this synthetic task rather than on standard benchmarks because it isolates retrieval from other capabilities. Benchmarks such as MMLU also depend on parametric knowledge, reasoning, and answer selection, which makes it hard to tell whether an intervention affects retrieval specifically. Our task fixes the relevant information inside the prompt, places the target at a known position, and lets us build matched prompts that differ only in the queried key, properties that the interventions of Section~\ref{sec:circuit} require.

\paragraph{Models.}
We study two families of hybrid language models with representative linear-attention mechanisms. The Qwen3.5 family includes \qsmall{} (0.8B parameters, 24 layers) and \qlarge{} (9B parameters, 32 layers), both using Gated DeltaNet with a 3:1 ratio of linear-attention to softmax-attention layers. The Granite-4.0-H family includes \gsmall{} (1B parameters) and \glarge{} (7B total parameters, approximately 1B active per token), both using Mamba-2 with four softmax-attention layers among 40 layers. \glarge{} is a mixture-of-experts model. Section~\ref{sec:scaling} extends our analysis to Qwen3.5-27B, which contains 48 linear-attention and 16 softmax-attention layers (3:1 ratio). We define a \emph{block} as a maximal sequence of consecutive linear-attention layers. Main-text figures focus on \qlarge{}, with results for the other models reported in Appendix~\ref{sec:additional-models}.

\subsection{Retrieval Emerges Abruptly at a Softmax Layer}
\label{sec:pruning}

\begin{figure}[t]
    \centering
    \includegraphics[width=\textwidth]{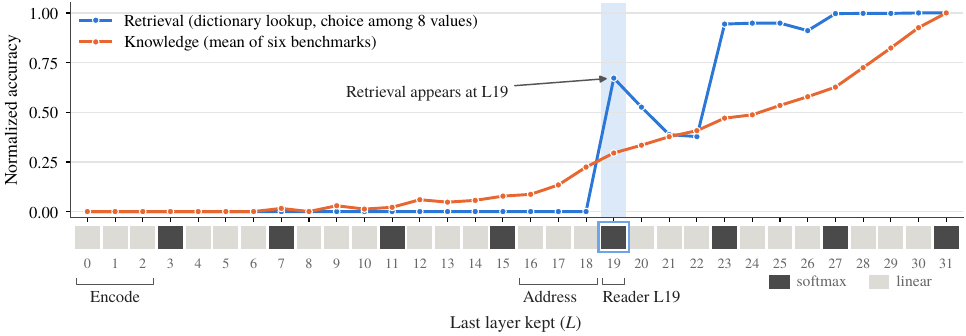}
    \caption{Performance of \qlarge{} truncated after layer $L$.  Retrieval appears abruptly when the softmax layer L19 is kept; knowledge grows gradually with depth.}
    \label{fig:pruning}
\end{figure}

To locate where retrieval is completed, we follow the progressive-pruning analysis of \citep{gna2025}. We truncate the network after layer $L$: the pruned model keeps layers $0,\dots,L$, removes every later layer (both its token mixer and its MLP), and feeds the remaining hidden state directly into the final normalization and language-model (LM) head. As a contrast, we also evaluate knowledge-focused performance on six benchmarks (ARC-Challenge, ARC-Easy, PIQA, WinoGrande, OpenBookQA, and HellaSwag) using zero-shot evaluation with lm-evaluation-harness \citep{gao2024lmeval} and averaged over the six tasks. A truncated model rarely ranks the answer first over the full vocabulary, so we score retrieval in the same multiple-choice manner: the prediction is the dictionary value with the highest logit, and random guessing yields $\mathrm{chance}=1/8$. We report normalized accuracy, $(\mathrm{acc}-\mathrm{chance})/(\mathrm{acc}_{\mathrm{full}}-\mathrm{chance})$, which maps random-guessing accuracy to 0 and the unpruned model's accuracy to 1.

Figure~\ref{fig:pruning} shows the results for \qlarge{} and reveals three key observations:
\begin{enumerate}
    \item \textbf{Knowledge develops gradually, while retrieval emerges abruptly.}
    Knowledge-focused performance improves steadily as more layers are retained. Retrieval remains at the level of random guessing until layer 19, regardless of how many preceding linear-attention layers are retained.

    \item \textbf{A single softmax layer marks the onset of retrieval.}
    Retaining the softmax layer L19 raises normalized retrieval accuracy from 0 to 0.67. Retrieval accuracy may fluctuate between softmax layers, possibly because intermediate representations are not decodable by the final LM head. This does not affect where retrieval first appears.\looseness-1

    \item \textbf{Retrieval requires only part of the network.}
    Retaining 24 of 32 layers preserves 0.94 of the full model's normalized retrieval accuracy, compared with only 0.47 of its knowledge-focused performance.
\end{enumerate}

\section{The Encode--Address Circuit}
\label{sec:circuit}

Progressive pruning localizes retrieval to a softmax layer but does not reveal what the preceding linear-attention layers contribute, since all layers below the reader remain intact. To retrieve the correct value, the reader must match its query against the stored keys and aggregate their values. We therefore ask how the preceding linear-attention layers enable this lookup.

We use two complementary interventions on linear-attention blocks, i.e., consecutive layers. \emph{Removal} tests whether a block is \emph{necessary} for retrieval; \emph{transplant} tests whether its output is \emph{sufficient} to determine what is retrieved. Unlike truncation, both interventions target only the linear-attention token mixers, leaving the MLPs intact, because our question concerns what linear attention computes.

We intervene at the block level for two reasons. First, pruning shows that retrieval changes primarily at softmax-layer boundaries, making blocks a natural unit of analysis. Second, single-layer interventions can miss effects that emerge across multiple layers: removing or transplanting individual linear-attention layers has little effect, except for removing the first layer (Appendix~\ref{app:single-layer}). We examine individual layers when necessary, as in the Encode block.

\begin{figure}[t]
    \centering
    \includegraphics[width=\textwidth]{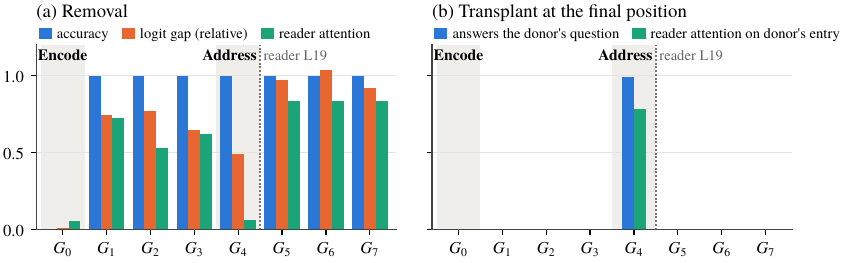}
    \caption{Removal and transplant of each linear-attention block of \qlarge{} (\ttt{chat8}); $G_k$ comprises layers $4k$ to $4k{+}2$. (a) Removing the Encode block $G_0$ destroys retrieval. Removing the Address block $G_4$ leaves accuracy at 1.00 but halves the logit gap and reduces the reader's attention on the queried entry from 0.83 to 0.06. (b) Transplanting $G_4$'s output at the final position from a prompt about another key makes the model answer that key and moves the reader to the donor's entry; no other block has an effect.}
    \label{fig:circuit}
\end{figure}

Figure~\ref{fig:circuit} presents the removal and transplant results of Qwen3.5-9B on \ttt{chat8}. It has eight blocks, $G_0$ to $G_7$, of three GDN layers each; $G_4$ (layers 16--18) immediately precedes the reader layer L19.  Additional models and prompt formats are reported in Appendix~\ref{sec:additional-models}. Together, these interventions reveal two distinct roles for linear attention: an \emph{Encode} block that re-encodes token identity and an \emph{Address} block that determines which entry the reader retrieves.

\subsection{Removal: Identifying Critical Blocks}
\label{sec:removal}

\emph{Removal} sets a block's token-mixer output to zero at every position, leaving its MLPs and all softmax layers intact. We evaluate its effect using three metrics. \emph{Accuracy} measures whether the model predicts the correct value. The \emph{logit gap} measures the model's preference for the correct value over an alternative (e.g., ``yellow'' over ``green'' for ``banana=yellow''). We swap the two values to create a paired prompt and average the correct-minus-alternative logit differences across both prompts. A larger positive gap indicates a stronger preference for the correct value. \emph{Reader attention} measures the attention mass assigned by the reader, from the final position, to the queried dictionary entry. A higher value indicates stronger attention to the correct entry.

Figure~\ref{fig:circuit}a shows three patterns. First, removing the earliest block, $G_0$, nearly eliminates retrieval (accuracy 0.01, normalized logit gap 0.01, reader attention 0.06). Removing its first layer alone reduces accuracy to 0.05. Second, removing $G_4$, immediately before the reader, preserves accuracy at 1.00 but reduces the logit gap from $+23.8$ to $+11.6$ and reader attention from 0.83 to 0.06. Thus, although the model still predicts the correct answer, its preference for that answer weakens substantially, and the identified reader no longer attends reliably to the target. Third, removing intermediate blocks $G_1$--$G_3$ partially weakens reader attention (0.53--0.73), while blocks after the reader have little effect.\looseness-1

The result for $G_4$ is particularly revealing. Its removal disrupts the reader at L19, but later softmax layers (L23 and L27) compensate and recover the answer, as discussed in Section~\ref{sec:scaling}. In models without such redundancy, removing the block before the reader reduces accuracy to 0.13 in \qsmall{} and 0.00 in \gsmall{}. These results establish that $G_0$ is essential for retrieval, while $G_4$ is critical to the reader but not necessarily to the final prediction. Removal alone therefore cannot establish which block determines what the model retrieves.

In summary, the earliest Encode block is essential for retrieval, while the Address block immediately before the reader is critical to its lookup but may be dispensable for the final prediction due to redundancy in later softmax layers.

\subsection{Transplant: Redirecting Retrieval}
\label{sec:transplant}

\emph{Transplant} tests whether a block's output is sufficient to determine what is retrieved. We construct a donor prompt A and a host prompt B with the same dictionary but different queried keys (e.g., A asks about ``banana'' and B about ``ginger''). Since their lengths and dictionary positions are identical, we can replace a block's mixer output at the final position in B with the corresponding output from A. If the transplanted representation determines the retrieval target, B should now return A's value instead of its own. We measure the success rate over 100 donor--host pairs.


Figure~\ref{fig:circuit}b shows that transplanting $G_4$ redirects the model to the donor's answer in 0.99 of the pairs. The reader's attention also shifts to the donor's entry. In contrast, transplanting any other block, including $G_0$, has no effect (0.00). Transplanting $G_4$ over the dictionary positions rather than the final position also fails. This result holds across prompt formats (0.97 on \ttt{list8}, 0.98 on \ttt{rev8}). Thus, the output of a single linear-attention block at the final position determines which entry the reader retrieves. We call this the \emph{Address} block.\looseness-1


Removal and transplant reveal a striking distinction. Across all prompt formats, removing the Address block preserves accuracy at 0.76--1.00 but roughly halves the logit gap, whereas transplanting its output redirects retrieval in 0.97--0.99 of cases. Together, these results establish the central mechanism of the Encode--Address circuit: the Encode block enables retrieval, while the Address block determines which entry the softmax reader retrieves.

\subsection{The Role of Encode}
\label{sec:encode}

Removing $G_0$ destroys retrieval, yet transplanting its output at the final position has no effect. To understand its role, we perform two token-level interventions. In a \emph{value swap}, we change a dictionary entry from "apple=red'' to "apple=yellow'', then replace the Encode output for yellow'' with that recorded for red'' in the original prompt. The intervention succeeds if the model answers red'' rather than yellow'' when queried about ``apple''. In a \emph{key swap}, we similarly exchange the keys of two entries and restore their original Encode outputs. As a control, we also replace the input embeddings at the swapped positions with those of the original tokens. For all interventions, success means recovering the original key--value associations.

\begin{table}[t]
\centering
\caption{The Encode block of \qlarge{} (\ttt{chat8}). Left: success rate of token-level interventions in restoring the original key--value associations. Right: retrieval accuracy and WikiText-2 perplexity under modifications of $G_0$.}
\label{tab:encode}
\small
\begin{tabular}{lcc}
\toprule
Patched at swapped tokens & Value swap & Key swap \\
\midrule
Encode block output & 0.98 & 0.98 \\
Input token embedding & \textbf{1.00} & \textbf{1.00} \\
Any other block's output & 0.00 & 0.00 \\
No Patching & 0.00 & 0.00 \\
\bottomrule
\end{tabular}
\hspace{1.5em}
\begin{tabular}{lcc}
\toprule
$G_0$ & Accuracy & Perplexity \\
\midrule
Intact & 1.00 & 10.6 \\
Token-wise linear projection & 1.00 & 44.7 \\
Mixers zeroed (MLPs kept) & 0.00 & 1059 \\
Removed & 0.00 & $6\times10^{4}$ \\
\bottomrule
\end{tabular}
\end{table}

Patching the Encode outputs restores the original associations in 0.98 of cases for both \emph{value} and \emph{key} swaps (Table~\ref{tab:encode}, left). Surprisingly, patching only the input token embeddings achieves 1.00, while patching any other block's output has no effect. This suggests that the information needed to restore retrieval is already present in the input embeddings, rather than being uniquely computed by Encode.

We further isolate the role of Encode by replacing $G_0$ with a token-wise linear projection fitted to its original inputs and outputs using ridge regression. This replacement preserves retrieval accuracy (1.00) but increases WikiText-2 perplexity from 10.6 to 44.7. These results suggest that Encode's role in \emph{retrieval} can be fulfilled by a simple linear transformation, although the original block performs additional computations important for general language modeling. Having established Encode's role in preserving token identity, we next investigate how Address constructs the representation that directs the softmax reader toward the queried key.

\section{How Is the Address Computed?}
\label{sec:construction}

Section~\ref{sec:transplant} establishes that the Address block's final-position output determines which dictionary entry the softmax reader retrieves. We now investigate how this output is constructed: whether it controls the reader's query or keys, where the retrieval target is stored, and how other blocks contribute to its computation.

\begin{figure}[t]
    \centering
    \includegraphics[width=\textwidth]{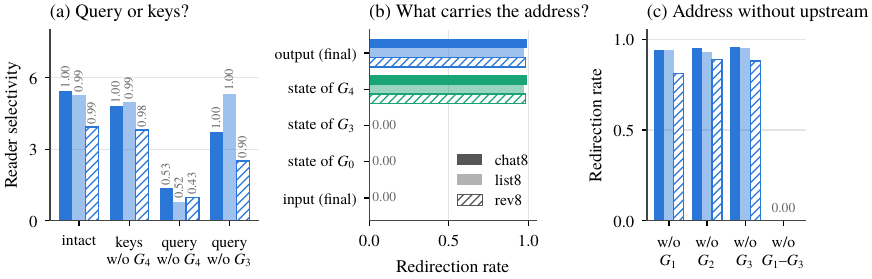}
    \caption{How the Address block constructs the retrieval target in \qlarge{}. (a) Reader selectivity when its query or keys are computed without the indicated block; numbers show top-1 rates. Solid, light, and hatched bars denote the three prompt formats. (b) Transplanting the Address block's final-position output, or its recurrent state, redirects retrieval; the states of earlier blocks and the block's input at the final position do not. (c) Address block outputs computed without any single upstream block still redirect retrieval, but those computed without all three do not.}
    \label{fig:construction}
\end{figure}

\subsection{The Address Block Determines the Reader's Query}
\label{sec:query-key}
A softmax head retrieves an entry by matching its final-position query against the dictionary keys. To determine whether the Address block controls the query or the keys, we run \qlarge{} with and without $G_4$, then recompute attention using (1) the query without $G_4$ and the original keys, or (2) the original query and the keys without $G_4$ (Figure~\ref{fig:construction}a). We measure \emph{selectivity}, the difference between the log attention mass on the queried entry and the mean log attention mass on other entries, and the \emph{top-1 rate}, the fraction of prompts where the queried entry receives the most attention. Higher selectivity indicates a stronger preference for the queried entry, while a higher top-1 rate indicates more frequent correct selection.

Figure~\ref{fig:construction}a shows that the Address block primarily controls the query. Using a query computed without $G_4$ reduces selectivity from 5.45 to 1.36, with the reader selecting the correct entry on only about half of the prompts. In contrast, using keys computed without $G_4$ leaves selectivity nearly unchanged (4.80) and the top-1 rate at 1.00. Removing the preceding block $G_3$ only mildly weakens query selectivity without changing the reader's choice. Thus, these results indicate that the Address block primarily determines the reader's attention \emph{query}, rather than its keys.

\subsection{The Recurrent State Determines the Retrieval Target}
\label{sec:state}

The preceding results establish that the Address block determines the reader's attention query. We next investigate how its recurrent state contributes to this computation as plotted in Figure~\ref{fig:construction}b.

As shown in Section~\ref{sec:transplant}, transplanting the Address block's final-position output redirects retrieval to the donor's entry in nearly all cases. To determine whether its recurrent state carries the information needed to produce this output, we run donor and host prompts up to the penultimate token, copy the donor's recurrent states in all three Address layers ($G_4$) into the host, and let the host process its own final token. This also redirects retrieval in 0.99 of cases. In contrast, transplanting the recurrent states of $G_0$ or $G_3$ has no effect (0.00).

As a further control, we replace the host's input to $G_4$ at the final position with the donor's, while keeping the host's recurrent state unchanged. This also fails to redirect retrieval (0.00). Together, these results indicate that the Address block's recurrent state carries the information determining the retrieval target, which the block uses to construct its final-position output. The softmax reader then uses this output to form the attention query that selects the corresponding dictionary entry.



\section{Does the Address Transfer Across Architectures?}
\label{sec:transfer}

Qwen3.5 and Granite-4.0-H are trained independently and differ in their proportion of linear-attention layers, recurrent mechanisms (Gated DeltaNet vs. Mamba-2), and model sizes. Despite these architectural differences, can the Address block learned by one model support retrieval in the other? We investigate whether both its output and computation transfer across the two model families.

We test two pairs of similarly sized models: \qsmall{} with \gsmall{}, and \qlarge{} with \glarge{}. Since their hidden representations differ, we fit linear alignment maps between models using ridge regression on paired activations without updating either model's weights.

\begin{table*}[t]
\centering
\caption{Cross-architecture transfer on \ttt{chat8}.
\emph{Output transplantation}: success rate of redirecting
retrieval using the donor Address block's output, compared
with a same-model baseline using two prompts processed by
the receiver model.
\emph{Block replacement}: receiver accuracy after replacing
its Address block with the donor block or a control.
$^a$Both Address blocks of \glarge{} are replaced.
$^b$\qlarge{} is evaluated after L19 (intact accuracy: 0.75),
using both Address blocks of \glarge{} as the donor.}
\label{tab:transfer}

\small
\setlength{\tabcolsep}{3pt}
\begin{tabular}{lcccccc}
\toprule
& \multicolumn{2}{c}{Output transplant}
& \multicolumn{4}{c}{Block replacement} \\
\cmidrule(lr){2-3}
\cmidrule(lr){4-7}
Donor $\rightarrow$ receiver
& \makecell{Cross-\\model}
& \makecell{Same-\\model}
& Removed
& \makecell{Linear\\proj.}
& \makecell{Zeroed\\mixer}
& \makecell{\textbf{Donor}\\\textbf{Address}} \\
\midrule
\qsmall{} $\rightarrow$ \gsmall{}
& 0.98 & 1.00 & 0.01 & 0.24 & 0.28 & \textbf{0.92} \\
\gsmall{} $\rightarrow$ \qsmall{}
& 0.91 & 0.97 & 0.30 & 0.28 & 0.28 & \textbf{0.95} \\
\qlarge{} $\rightarrow$ \glarge{}$^{a}$
& 0.99 & 1.00 & 0.00 & 0.14 & 0.18 & \textbf{0.94} \\
\glarge{} $\rightarrow$ \qlarge{}$^{b}$
& 0.83 & 0.92 & 0.35 & 0.30 & 0.21 & 0.46 \\
\bottomrule
\end{tabular}
\end{table*}

\subsection{The Address Block's Output Transfers}
\label{sec:output-transfer}
We first test whether one model can use the Address block output computed by another. Donor and receiver process the same dictionary but ask about different keys. We transplant the donor's final-position Address block output into the receiver and measure how often the receiver answers the donor's question. As a baseline, we also transplant outputs between two prompts processed by the same receiver model, reported in the Same-model column of Table~\ref{tab:transfer}.

For the smaller models, transplantation succeeds in both directions: 0.98 from \qsmall{} to \gsmall{} and 0.91 in reverse on \ttt{chat8}, close to their same-model baselines of 1.00 and 0.97, respectively. Success rates range from 0.87 to 1.00 across formats, and the receiver's softmax reader shifts its attention to the donor's entry.

The larger pair introduces an additional complication. In \glarge{}, retrieval depends on two reader layers, L25 and L35, preceded by Address blocks $G_2$ and $G_3$, respectively. Transplanting $G_2$ alone redirects the first reader but rarely changes the final answer (0.03--0.06), whereas transplanting both blocks redirects both readers and the answer (1.00, 1.00, and 0.90 across formats). Thus, transplantation must account for both Address blocks in \glarge{}.

Accordingly, transplanting the Address block output of \qlarge{} into both blocks of \glarge{} redirects retrieval in 0.99 of cases, matching the in-family result (1.00). Transplanting into only the first block rarely succeeds (0.04). In the reverse direction, the two Address block outputs of \glarge{} redirect \qlarge{} in 0.83 of cases, compared with 0.92 for in-family transplantation.

These results show that the Address block's output can guide retrieval across independently trained architectures, with cross-model transplantation approaching the success rates of same-model transplantation. 

\subsection{The Address Block's Computation Transfers}
\label{sec:computation-transfer}

Transplanting the Address block's output shows that the receiver can use a retrieval target computed by another model. We next test whether the donor's Address block can replace the receiver's computation itself.

We remove the receiver's Address block and insert the donor's block in its place, using linear maps to connect their hidden representations. Unlike output transplantation, the donor block now processes the receiver's sequence. We compare this replacement against three controls: (1) removing the receiver's Address block without replacement, (2) replacing it with a token-wise linear projection, and (3) inserting the donor's Address block but zeroing its linear-attention mixer outputs while retaining its MLPs. These controls test whether successful transfer requires the donor Address block's recurrent computation rather than a simple linear transformation or its remaining components.

For the smaller models, block replacement succeeds in both directions (Table~\ref{tab:transfer}). Replacing \gsmall{}'s Address block with that of \qsmall{} achieves 0.92 accuracy, compared with 0.01 when the receiver's block is removed without replacement. The reverse replacement achieves 0.95, compared with 0.30 without replacement. In contrast, a token-wise linear projection or a donor block with zeroed mixer outputs achieves at most 0.28. These results indicate that successful transfer depends on the donor Address block's recurrent token-mixing computation, rather than a simple linear transformation.

For the larger models, replacement is more challenging because \glarge{} uses two Address blocks, whereas \qlarge{} relies primarily on one (Section~\ref{sec:output-transfer}). We therefore replace both Address blocks of \glarge{} and evaluate \qlarge{} after L19 to prevent compensation by later readers. Replacing both blocks of \glarge{} with \qlarge{}'s Address block achieves 0.94 accuracy, compared with 0.00 without replacement, 0.14 with a token-wise linear projection, and 0.18 with zeroed donor mixers.

The reverse transfer is less effective. Replacing \qlarge{}'s Address block with both Address blocks of \glarge{}, run sequentially without their intervening softmax layer, achieves 0.46 accuracy. This exceeds the removal control (0.35) and either donor block alone (0.34--0.36), but remains below the intact receiver's accuracy (0.75). This partial recovery suggests that computation distributed across multiple Address blocks plus a softmax layer is harder to transfer.

Together, these experiments show that the Address block's computation can transfer between Gated DeltaNet and Mamba-2, particularly when donor and receiver have compatible Address block structures.

\section{Scaling: From One Address Block to Several}
\label{sec:scaling}

We examine how Address computation changes across six Qwen3.5 models: five dense models from \qsmall{} to Qwen3.5-27B, and the mixture-of-experts Qwen3.5-35B-A3B (3B active parameters). All six place one softmax layer after every three linear-attention layers. We investigate how many Address blocks are needed to redirect retrieval, which softmax readers they guide, and whether retrieval depends on them (Table~\ref{tab:scaling}).

\begin{table*}[t]
\centering
\caption{Address blocks and the readers they guide in Qwen3.5 models. \emph{Removal}: accuracy after removing the Address blocks. \emph{Transplant}: fraction of prompts redirected to the donor's target when the Address blocks are transplanted jointly.}
\label{tab:scaling}
\small
\begin{tabular}{@{}lcccc@{}}
\toprule
Model & Linear / softmax & Address block $\to$ guided readers & Removal & Transplant \\
\midrule
\qsmall{} & 18 / 6 & $G_3 \to$ L15, L19, L23 & 0.06 & 0.99 \\
Qwen3.5-2B & 18 / 6 & $G_3 \to$ L15, L19, L23 & 0.41 & 1.00 \\
Qwen3.5-4B & 24 / 8 & $G_4 \to$ L19, L23, L27 & 0.94 & 0.70 \\
\qlarge{} & 24 / 8 & $G_4 \to$ L19, L23, L27 & 0.92 & 0.98 \\
\midrule
Qwen3.5-27B & 48 / 16 & $G_8$--$G_{10} \to$ L39, L43; $G_{11}, G_{12} \to$ L51, L59 & 1.00 & 0.98 \\
Qwen3.5-35B-A3B & 30 / 10 & $G_6 \to$ L27; $G_7 \to$ L31; $G_8 \to$ L35 & 0.34 & 0.75 \\
\bottomrule
\end{tabular}
\end{table*}

\paragraph{One Address block guides multiple readers up to 9B.}
The four smaller models share a common structure: a single Address block immediately before the first main reader guides all three readers. Transplanting this block's final-position output redirects retrieval in 0.70--1.00 of cases for these models. 

However, the necessity of the Address block differs across model sizes. Removing it reduces retrieval accuracy to 0.06 at 0.8B and 0.41 at 2B, but leaves accuracy at 0.94 and 0.92 at 4B and 9B. In the larger models with higher redundancy, the linear-attention blocks after the Address block $G_4$ also carry a weaker signal of which entry is queried.  When $G_4$ is present, its stronger signal dominates and all readers follow it; when it is removed, the weaker signal is still enough for the later readers to retrieve the correct value.

\paragraph{Address computation becomes distributed in larger models.}
Multiple Address blocks emerge in larger models. In Qwen3.5-27B and Qwen3.5-35B-A3B, no single block can redirect retrieval. Instead, different Address blocks guide their own readers (Appendix~\ref{app:scaling}), and jointly transplanting these blocks redirects 0.98 and 0.75 of answers, respectively.

How Address blocks guide readers also changes with scale. In smaller models (up to 9B), the first reader influences subsequent readers, allowing a single Address block to guide multiple readers. In larger models (above 9B), each Address block guides a different set of readers, providing more independent pathways for retrieval. However, their necessity differs between Qwen3.5-27B and Qwen3.5-35B-A3B. In the dense 27B model, removing the Address blocks leaves retrieval accuracy at 1.00. In contrast, removing the Address blocks reduces accuracy to 0.34 in the MoE 35B-A3B model. We leave a systematic comparison of dense and MoE architectures for future work.

In summary, the Address computation becomes more distributed in larger models, with multiple blocks contributing to the final retrieval query. This distribution allows for redundancy and robustness in retrieval.

\section{Conclusion}


In hybrid language models, softmax attention performs retrieval, but linear attention helps determine what is retrieved. We identify an Encode--Address circuit in which the Address block carries the retrieval target in its recurrent state and constructs the final-position representation from which the softmax reader forms its attention query. This representation causally redirects retrieval even when removing the Address block leaves accuracy intact. Both the representation and its underlying computation transfer across Gated DeltaNet and Mamba-2, while larger models distribute Address computation across multiple blocks. Together, these findings reveal a division of labor: linear attention constructs the query, while softmax attention performs the lookup. Thus, linear attention serves not only as compressed memory but also as an active component in determining what the model retrieves.

\bibliographystyle{plainnat}
\bibliography{references}

\begin{thebibliography}{23}
\providecommand{\natexlab}[1]{#1}
\providecommand{\url}[1]{\texttt{#1}}
\expandafter\ifx\csname urlstyle\endcsname\relax
  \providecommand{\doi}[1]{doi: #1}\else
  \providecommand{\doi}{doi: \begingroup \urlstyle{rm}\Url}\fi

\bibitem[Arora et~al.(2024{\natexlab{a}})Arora, Eyuboglu, Timalsina, Johnson,
  Poli, Zou, Rudra, and R{\'e}]{arora2024zoology}
Simran Arora, Sabri Eyuboglu, Aman Timalsina, Isys Johnson, Michael Poli, James
  Zou, Atri Rudra, and Christopher R{\'e}.
\newblock Zoology: Measuring and improving recall in efficient language models.
\newblock In \emph{International Conference on Learning Representations
  (ICLR)}, 2024{\natexlab{a}}.

\bibitem[Arora et~al.(2024{\natexlab{b}})Arora, Eyuboglu, Zhang, Timalsina,
  Alberti, Zinsley, Zou, Rudra, and R{\'e}]{arora2024based}
Simran Arora, Sabri Eyuboglu, Michael Zhang, Aman Timalsina, Silas Alberti,
  Dylan Zinsley, James Zou, Atri Rudra, and Christopher R{\'e}.
\newblock Simple linear attention language models balance the recall-throughput
  tradeoff.
\newblock In \emph{Proceedings of the 41st International Conference on Machine
  Learning (ICML)}, 2024{\natexlab{b}}.

\bibitem[Bick et~al.(2025)Bick, Xing, and Gu]{gna2025}
Aviv Bick, Eric Xing, and Albert Gu.
\newblock Understanding the skill gap in recurrent language models: The role of
  the gather-and-aggregate mechanism.
\newblock In \emph{Proceedings of the 42nd International Conference on Machine
  Learning (ICML)}, volume 267 of \emph{Proceedings of Machine Learning
  Research}, pages 4324--4344, 2025.
\newblock Code: \url{https://github.com/goombalab/Gather-and-Aggregate}.

\bibitem[Dao and Gu(2024)]{dao2024mamba2}
Tri Dao and Albert Gu.
\newblock Transformers are {SSM}s: Generalized models and efficient algorithms
  through structured state space duality.
\newblock In \emph{Proceedings of the 41st International Conference on Machine
  Learning (ICML)}, 2024.

\bibitem[Endy et~al.(2025)Endy, Grosbard, Ran-Milo, Slutzky, Tshuva, and
  Giryes]{endy2025knockout}
Nir Endy, Idan~Daniel Grosbard, Yuval Ran-Milo, Yonatan Slutzky, Itay Tshuva,
  and Raja Giryes.
\newblock {Mamba} knockout for unraveling factual information flow.
\newblock In \emph{Proceedings of the 63rd Annual Meeting of the Association
  for Computational Linguistics (Volume 1: Long Papers)}, 2025.

\bibitem[Gao et~al.(2024)Gao, Tow, Abbasi, Biderman, Black, DiPofi, Foster,
  Golding, Hsu, Le~Noac'h, Li, McDonell, Muennighoff, Ociepa, Phang, Reynolds,
  Schoelkopf, Skowron, Sutawika, Tang, Thite, Wang, Wang, and
  Zou]{gao2024lmeval}
Leo Gao, Jonathan Tow, Baber Abbasi, Stella Biderman, Sid Black, Anthony
  DiPofi, Charles Foster, Laurence Golding, Jeffrey Hsu, Alain Le~Noac'h,
  Haonan Li, Kyle McDonell, Niklas Muennighoff, Chris Ociepa, Jason Phang,
  Laria Reynolds, Hailey Schoelkopf, Aviya Skowron, Lintang Sutawika, Eric
  Tang, Anish Thite, Ben Wang, Kevin Wang, and Andy Zou.
\newblock The language model evaluation harness.
\newblock \url{https://github.com/EleutherAI/lm-evaluation-harness}, 2024.

\bibitem[{IBM}(2025)]{granite4}
{IBM}.
\newblock {IBM Granite 4.0}: Hyper-efficient, high performance hybrid models
  for enterprise.
\newblock
  \url{https://www.ibm.com/new/announcements/ibm-granite-4-0-hyper-efficient-high-performance-hybrid-models},
  October 2025.

\bibitem[Jelassi et~al.(2024)Jelassi, Brandfonbrener, Kakade, and
  Malach]{jelassi2024repeat}
Samy Jelassi, David Brandfonbrener, Sham~M. Kakade, and Eran Malach.
\newblock Repeat after me: Transformers are better than state space models at
  copying.
\newblock In \emph{Proceedings of the 41st International Conference on Machine
  Learning (ICML)}, 2024.

\bibitem[{Kimi Team}(2025)]{kimi2025linear}
{Kimi Team}.
\newblock {Kimi Linear}: An expressive, efficient attention architecture.
\newblock \emph{arXiv preprint arXiv:2510.26692}, 2025.

\bibitem[Lenz et~al.(2025)Lenz, Lieber, Arazi, et~al.]{lieber2024jamba}
Barak Lenz, Opher Lieber, Alan Arazi, et~al.
\newblock {Jamba}: Hybrid transformer-{Mamba} language models.
\newblock In \emph{International Conference on Learning Representations
  (ICLR)}, 2025.

\bibitem[Lieberum et~al.(2023)Lieberum, Rahtz, Kram{\'a}r, Nanda, Irving, Shah,
  and Mikulik]{lieberum2023mcq}
Tom Lieberum, Matthew Rahtz, J{\'a}nos Kram{\'a}r, Neel Nanda, Geoffrey Irving,
  Rohin Shah, and Vladimir Mikulik.
\newblock Does circuit analysis interpretability scale? evidence from multiple
  choice capabilities in {Chinchilla}.
\newblock \emph{arXiv preprint arXiv:2307.09458}, 2023.

\bibitem[{NVIDIA}(2025)]{nvidia2025nemotronh}
{NVIDIA}.
\newblock {Nemotron-H}: A family of accurate and efficient hybrid
  {Mamba}-transformer models.
\newblock \emph{arXiv preprint arXiv:2504.03624}, 2025.

\bibitem[Olsson et~al.(2022)Olsson, Elhage, Nanda, Joseph, DasSarma, Henighan,
  Mann, Askell, Bai, Chen, et~al.]{olsson2022induction}
Catherine Olsson, Nelson Elhage, Neel Nanda, Nicholas Joseph, Nova DasSarma,
  Tom Henighan, Ben Mann, Amanda Askell, Yuntao Bai, Anna Chen, et~al.
\newblock In-context learning and induction heads.
\newblock \emph{Transformer Circuits Thread}, 2022.
\newblock URL
  \url{https://transformer-circuits.pub/2022/in-context-learning-and-induction-heads/index.html}.

\bibitem[Paulo et~al.(2025)Paulo, Marshall, and Belrose]{paulo2024transfer}
Gon{\c{c}}alo Paulo, Thomas Marshall, and Nora Belrose.
\newblock Do transformer interpretability methods transfer to {RNNs}?
\newblock In \emph{Proceedings of the AAAI Conference on Artificial
  Intelligence}, volume~39, pages 27565--27572, 2025.

\bibitem[{Qwen Team}(2026)]{qwen35}
{Qwen Team}.
\newblock {Qwen3.5}: Towards native multimodal agents, February 2026.
\newblock URL \url{https://qwen.ai/blog?id=qwen3.5}.

\bibitem[Ren et~al.(2025)Ren, Liu, Lu, Shen, Liang, and Chen]{ren2024samba}
Liliang Ren, Yang Liu, Yadong Lu, Yelong Shen, Chen Liang, and Weizhu Chen.
\newblock {Samba}: Simple hybrid state space models for efficient unlimited
  context language modeling.
\newblock In \emph{International Conference on Learning Representations
  (ICLR)}, 2025.

\bibitem[Schlag et~al.(2021)Schlag, Irie, and
  Schmidhuber]{schlag2021fastweights}
Imanol Schlag, Kazuki Irie, and J{\"u}rgen Schmidhuber.
\newblock Linear transformers are secretly fast weight programmers.
\newblock In \emph{Proceedings of the 38th International Conference on Machine
  Learning (ICML)}, 2021.

\bibitem[Sen~Sharma et~al.(2024)Sen~Sharma, Atkinson, and Bau]{sharma2024mamba}
Arnab Sen~Sharma, David Atkinson, and David Bau.
\newblock Locating and editing factual associations in {Mamba}.
\newblock In \emph{Conference on Language Modeling (COLM)}, 2024.

\bibitem[Waleffe et~al.(2024)Waleffe, Byeon, Riach, Norick, Korthikanti, Dao,
  Gu, Hatamizadeh, Singh, Narayanan, Kulshreshtha, Singh, Casper, Kautz,
  Shoeybi, and Catanzaro]{waleffe2024empirical}
Roger Waleffe, Wonmin Byeon, Duncan Riach, Brandon Norick, Vijay Korthikanti,
  Tri Dao, Albert Gu, Ali Hatamizadeh, Sudhakar Singh, Deepak Narayanan, Garvit
  Kulshreshtha, Vartika Singh, Jared Casper, Jan Kautz, Mohammad Shoeybi, and
  Bryan Catanzaro.
\newblock An empirical study of {Mamba}-based language models.
\newblock \emph{arXiv preprint arXiv:2406.07887}, 2024.

\bibitem[Wang et~al.(2024)Wang, Paliotta, May, Rush, and
  Dao]{wang2024mamballama}
Junxiong Wang, Daniele Paliotta, Avner May, Alexander~M. Rush, and Tri Dao.
\newblock The {Mamba} in the {Llama}: Distilling and accelerating hybrid
  models.
\newblock In \emph{Advances in Neural Information Processing Systems
  (NeurIPS)}, 2024.

\bibitem[Wen et~al.(2025)Wen, Dang, and Lyu]{wen2024rnns}
Kaiyue Wen, Xingyu Dang, and Kaifeng Lyu.
\newblock {RNNs} are not transformers (yet): The key bottleneck on in-context
  retrieval.
\newblock In \emph{International Conference on Learning Representations
  (ICLR)}, 2025.

\bibitem[Wu et~al.(2025)Wu, Wang, Xiao, Peng, and Fu]{wu2024retrieval}
Wenhao Wu, Yizhong Wang, Guangxuan Xiao, Hao Peng, and Yao Fu.
\newblock Retrieval head mechanistically explains long-context factuality.
\newblock In \emph{International Conference on Learning Representations
  (ICLR)}, 2025.

\bibitem[Yang et~al.(2025)Yang, Kautz, and Hatamizadeh]{yang2024gdn}
Songlin Yang, Jan Kautz, and Ali Hatamizadeh.
\newblock Gated delta networks: Improving {Mamba2} with delta rule.
\newblock In \emph{International Conference on Learning Representations
  (ICLR)}, 2025.

\end{thebibliography}

\clearpage
\appendix
\begin{center}
{\LARGE\bfseries Appendix}
\end{center}
\vspace{0.5em}
\section{Reproduction}
\label{app:repro}

All experiments run against frozen public checkpoints with eager attention. Interventions are forward hooks on the token-mixer outputs; no weights are modified. Per-model result files, logs and prompts accompany the code.

\paragraph{Task generation.} Keys are drawn from 22 words and values from 20 colour/material words,
all single tokens in context, sampled without replacement. For each item the corrupted twin swaps
the queried value with a distractor's. Token positions are recovered from character spans through
the tokenizer's offset mapping, and items are batched only when their length and all entry
positions agree, so an intervention applies to identical positions across a batch. Items are
rejected unless clean and corrupted prompts have equal token length, differ only inside the target
and distractor entries, and both answers are single tokens.

\paragraph{Chat templates.} Qwen prompts use the model's template with an empty think block, so the
answer is the next token. For other families the wrapper is derived from the tokenizer's own chat
template; Granite's inserts a default system prompt, which is why its prompts are $\approx17$ tokens
longer for the same dictionary.

\section{Single-Layer versus Block Interventions}
\label{app:single-layer}

Table~\ref{tab:single-layer} compares interventions on the whole block that precedes the reader with the same interventions on each of its layers (\ttt{chat8}, final-position transplant). In both Qwen models no single layer changes the answer when transplanted, while the whole block does; \gsmall{} concentrates most of the effect in L23. Removing layer 0 alone breaks retrieval in every model, so we report the Encode block at layer resolution.

\begin{table}[H]
\centering
\caption{Whole-block versus single-layer interventions on the block preceding the reader (\ttt{chat8}). \emph{Transplant}: fraction of donor--host pairs that return the donor's value, with the reader's attention on the donor's entry in parentheses (host baseline $\le 0.01$); for single layers, the best layer of the block is shown. \emph{Removal}: accuracy with the block, or the most damaging single layer of the block, removed. The last column gives accuracy with layer 0 alone removed.}
\label{tab:single-layer}
\small
\begin{tabular}{llcccc|c}
\toprule
 & & \multicolumn{2}{c}{Transplant} & \multicolumn{2}{c|}{Removal} & Layer 0 \\
Model & Block & whole block & best single layer & whole block & worst single layer & removed \\
\midrule
\qsmall{} & L12--14 & \textbf{0.99} (0.84) & 0.00 (0.03, L14) & \textbf{0.13} & 0.90 (L14) & 0.00 \\
\qlarge{} & L16--18 & \textbf{0.99} (0.78) & 0.00 (0.10, L18) & 1.00 & 0.99 (L17) & 0.05 \\
\gsmall{} & L16--24 & \textbf{1.00} (0.83) & 0.93 (0.70, L23) & \textbf{0.00} & 0.73 (L24) & 0.00 \\
\bottomrule
\end{tabular}
\end{table}

\section{Additional Models and Prompt Formats}
\label{sec:additional-models}

Table~\ref{tab:models} repeats the removal and transplant analysis of Section~\ref{sec:circuit} for \qsmall{}, \gsmall{} and \glarge{}, and for all three prompt formats. In every model, removing the Encode block destroys retrieval, transplanting the Address redirects retrieval in 0.90--1.00 of the pairs, and no other single block does. \glarge{} reads the answer at two softmax layers, L25 and L35, preceded by $G_2$ (layers 16--24) and $G_3$ (layers 26--34). Transplanting $G_2$ alone moves the L25 reader to the donor's entry (attention 0.80 on \ttt{chat8}) but changes the answer in only 0.03 / 0.06 / 0.05 of the pairs, because L35 still reads with the host's query. We therefore treat $G_2$ and $G_3$ jointly as the Address of \glarge{}: transplanting both redirects 1.00 / 1.00 / 0.90 of the pairs, the last matching the model's own accuracy on \ttt{rev8}, and removing both reduces accuracy to at most 0.06.

\begin{table}[H]
\centering
\caption{The circuit across models (\ttt{chat8} / \ttt{list8} / \ttt{rev8}). Removal: accuracy with the block removed (intact $\geq 0.90$). Transplant: fraction of pairs that return the donor's value when the block's output is transplanted at the final position. For \glarge{}, the Address comprises two blocks, $G_2$ and $G_3$, which are removed and transplanted jointly; transplanting $G_2$ alone gives 0.03 / 0.06 / 0.05.}
\label{tab:models}
\small
\setlength{\tabcolsep}{4pt}
\begin{tabular}{llcccc}
\toprule
 & & \multicolumn{2}{c}{Removal} & \multicolumn{2}{c}{Transplant} \\
\cmidrule(lr){3-4}\cmidrule(lr){5-6}
Model & Address & Encode & Address & Address & best other block \\
\midrule
\qsmall{} & L12--14 & 0.00 / 0.00 / 0.00 & 0.13 / 0.03 / 0.01 & \textbf{0.99 / 1.00 / 0.97} & 0.00 / 0.00 / 0.00 \\
\qlarge{} & L16--18 & 0.01 / 0.00 / 0.00 & 1.00 / 1.00 / 0.76 & \textbf{0.99 / 0.97 / 0.98} & 0.00 / 0.00 / 0.00 \\
\gsmall{} & L16--24 & 0.00 / 0.00 / 0.00 & 0.00 / 0.01 / 0.00 & \textbf{1.00 / 0.98 / 0.92} & 0.00 / 0.00 / 0.00 \\
\glarge{} & L16--24, L26--34 & 0.00 / 0.00 / 0.00 & 0.01 / 0.06 / 0.00 & \textbf{1.00 / 1.00 / 0.90} & 0.00 / 0.00 / 0.00 \\
\bottomrule
\end{tabular}
\end{table}

\section{Address Blocks and Readers in Larger Models}
\label{app:scaling}

\begin{figure}[H]
    \centering
    \includegraphics[width=0.78\textwidth]{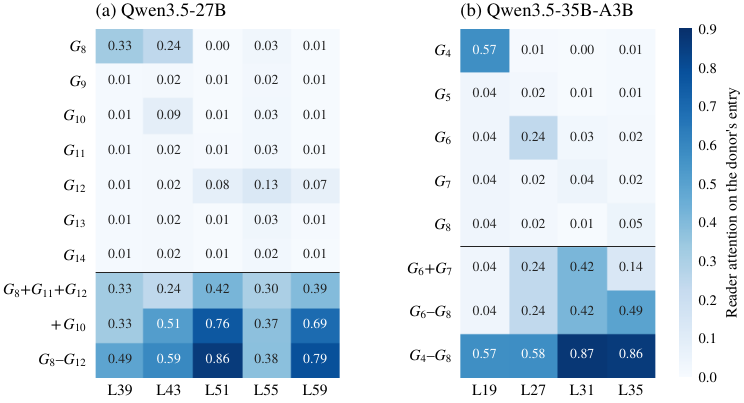}
    \caption{Attention of the top head of each reader layer on the donor's entry after transplanting single core blocks (above the line) or their combinations (below) at the final position, in Qwen3.5-27B (a) and Qwen3.5-35B-A3B (b). Intact host $\leq 0.04$.}
    \label{fig:scaling-readers}
\end{figure}

Figure~\ref{fig:scaling-readers} shows which readers each Address block guides in the two largest models: each block moves only the readers that follow it. At 35B-A3B, $G_6$ moves L27, $G_7$ moves L31 together with $G_6$, $G_8$ adds L35, and $G_4$ additionally moves L19. At 27B, $G_8$ moves L39 and L43, and L51 and L59 follow once $G_{11}$ and $G_{12}$ are added.

\end{document}